# Reservoir property image slices from the Groningen gas field for image translation and segmentation

Abdulrahman Al-Fakih[1,2*], Nabil Sariah[1], Ardiansyah Koeshidayatullah[1,2], and SanLinn I. Kaka[1,2]

[1]Department of Geosciences, King Fahd University of Petroleum and Minerals, Dhahran, 31261, Saudi Arabia.

[2]Center for Integrative Petroleum Research, King Fahd University of Petroleum and Minerals, Dhahran, 31261, Saudi Arabia

[*]Corresponding author: Abdulrahman Al-Fakih (alfakihabdulrahman2030@gmail.com)

***Reservoir characterization workflows increasingly rely on image-based and machine-learning/deep learning or even generative AI approaches, but openly available geological image datasets suitable for reproducible benchmarking remain limited. Here we describe a high-resolution dataset of reservoir-property image slices derived from the Groningen static geological model. The dataset contains aligned two-dimensional PNG images representing facies, porosity, permeability, and water saturation, generated from three-dimensional reservoir grids and prepared for downstream visualization, segmentation, and image-to-image translation tasks. In addition to the deposited original image corpus, we provide an archived software workflow for reproducing augmentation, mask generation, paired-image construction, and example baseline experiments. The resource is designed to support benchmarking of geological image analysis methods and the study of cross-domain relationships among reservoir properties. By separating the fixed image dataset from the reproducible processing workflow, this work provides a transparent foundation for reuse in geoscience, reservoir modeling, and machine-learning applications***.

## Background & Summary

High-resolution image representations of subsurface properties are increasingly important for reservoir characterization, geoscientific visualization, and data-driven modelling workflows[3–6]. Conventional reservoir modelling has long relied on geostatistical approaches and simplifying assumptions regarding spatial continuity, depositional architecture, and property distributions[7–11]. Although these approaches remain central to subsurface modelling, openly accessible datasets that provide aligned geological-property images derived from realistic reservoir models and organized for reproducible benchmarking remain limited[12,13].

Here we describe GeoPix, a dataset of reservoir-property image slices derived from the static geological model of the Groningen gas field. The dataset contains aligned two-dimensional images of facies, porosity, permeability, and water saturation extracted from three-dimensional reservoir grids and prepared for reuse in image-based reservoir characterization, semantic segmentation, and supervised image-to-image translation workflows. GeoPix is distributed through two linked outputs: the original high-resolution image corpus is deposited as a dataset at Zenodo[1], whereas the workflow used to reproduce augmentation, mask generation, paired-image construction, and example baseline demonstrations is archived separately as software at Zenodo[2]. This structure separates the fixed deposited data product from the reusable computational workflow and supports transparent reuse and reproducibility.

Recent advances in deep learning, particularly generative adversarial networks and conditional image-translation frameworks, have expanded the scope of image-based geoscientific analysis[13,14]. Conditional GANs such as Pix2Pix have proven especially relevant for supervised cross-domain translation tasks in which one geological representation is learned from another[15]. In geoscience applications, these approaches have been used to generate facies realizations and to learn relationships among geological and petrophysical domains[16,17]. Their broader development, however, has been constrained by the limited availability of high-resolution, publicly accessible datasets specifically structured for such tasks[12].

Within this context, Pix2Geomodel should be understood as a related downstream application framework built on aligned reservoir-property imagery rather than as the dataset described in the present paper. Earlier and ongoing developments of the Pix2Geomodel workflow have explored property-to-property translation, multi-scale extensions, enhanced facies and property prediction, and spatial continuity validation using Pix2Pix-based conditional generative models[18-22]. These studies demonstrate the potential scientific value of aligned reservoir image datasets, while also highlighting the need for a clearly documented data resource that can be reused independently of any single downstream modeling framework. GeoPix is intended to meet that need by providing the underlying image dataset and reproducible processing workflow in a form suitable for broader reuse.

GeoPix was derived from the high-resolution Petrel geological model of the Groningen gas field, developed by Nederlandse Aardolie Maatschappij and distributed through EPOS-NL and the Yoda data publication platform at Utrecht University[23,24]. The source model provides a well-documented geological framework from which aligned two-dimensional slices can be systematically extracted from three-dimensional reservoir-property arrays[25]. In addition to porosity, permeability, and water saturation, the dataset includes independently generated facies images to support cross-domain analysis and supervised learning applications. By separating the curated dataset from downstream application workflows, GeoPix provides a reusable resource for benchmarking, visualization, segmentation, and image-translation studies in subsurface geoscience.

## Geological Context: The Groningen Gas Field

The Groningen gas field, located in the northern Netherlands, is one of Europe's largest and most extensively studied natural gas reservoirs[26-28]. Discovered in 1959 through the drilling of the Slochteren-1 well, the field has played a pivotal role in Dutch energy supply for over six decades[29]. It has also served as a benchmark for geological modeling, reservoir simulation, and geomechanical analysis due to its well-documented structural complexity and extensive dataset availability[30].

Geologically, the Groningen reservoir is hosted within the Permian-aged Rotliegend Formation, which is overlain by a thick Zechstein salt seal[29,30]. The reservoir interval is primarily composed of aeolian and fluvial sandstones, interbedded with siltstones and claystones that introduce vertical heterogeneity[27,32]. This heterogeneity, along with varying diagenetic overprints, significantly affects porosity and permeability distributions across the field. The structural architecture of Groningen is characterized by a dense network of over 1100 normal faults, many of which act as partial flow barriers and influence both production behavior and subsurface stress[26,33].

The field is stratigraphically subdivided into 12 zones and further divided into 175 reservoir layers, defined by lithostratigraphic markers such as the Lower and Upper Slochteren Sandstones, Ameland Claystone, and Ten Boer Claystone[27]. These divisions reflect key depositional and diagenetic transitions that are critical for reservoir modeling and geomechanical analysis. Over the decades, the static model of the Groningen field has evolved from a simple, well-based representation to a high-resolution, faulted 3D grid[34]. The Groningen Field Review 2012 (GFR2012) was a significant milestone in this evolution, introducing a 100 × 100 m grid resolution and integrating approximately 700 simulation-relevant faults. These refinements were driven by increasing concerns over production-induced seismicity and the need for more accurate geomechanical modeling [23,26].

In later updates, improvements included enhanced porosity modeling via seismic inversion, refined fault interpretation based on dense 3D seismic coverage, and the incorporation of aquifer zones beyond the main reservoir body[35,36]. These enhancements enabled more realistic simulations of pressure depletion and stress redistribution, which are essential for assessing seismic risk[37]. The Groningen field's geological setting, along with its extensive seismic, well, and production datasets, makes it an ideal candidate for testing data-driven approaches to subsurface property modeling, such as those enabled by cGAN-based image translation frameworks[38,39].

## Methods

**Input data.** Reservoir property models for porosity, permeability, and water saturation were obtained from the Petrel geological model of the Groningen gas field made available through EPOS-NL and related repository infrastructure[23,24]. These source data were exported in GSLIB format with retained cell indices to preserve spatial correspondence during downstream processing. Because a facies representation was not directly available in the source release, facies images were generated independently within Petrel using geological interpretation and reservoir-property context.

A comprehensive overview of the GeoPix v1 dataset preparation workflow is presented in (Fig. 1), which outlines the complete pipeline from initial data extraction to final dataset assembly. The process begins with exporting property models (porosity, permeability, water saturation) from the Petrel platform in GSLIB format. These files are then preprocessed, converted to CSV, cleaned, and reshaped into 3D arrays from which 2D high-resolution slices are extracted. Custom color mapping and image enhancements are applied to preserve geological features. Each slice is augmented through geometric transformations to increase dataset diversity. Automated mask generation supports segmentation tasks, while structured pairing of facies and reservoir properties enables property-to-property translation. The resulting dataset is stratified into training, validation, and testing sets, with consistent naming and directory structures to ensure reproducibility and ease of use.

The source Groningen model used in this work was obtained from the EPOS-NL/Yoda data publication by NAM, which is described in the source metadata as open access and licensed under Creative Commons Attribution 4.0. GeoPix constitutes a derived image product generated from that source publication and is redistributed with attribution to the original data record.

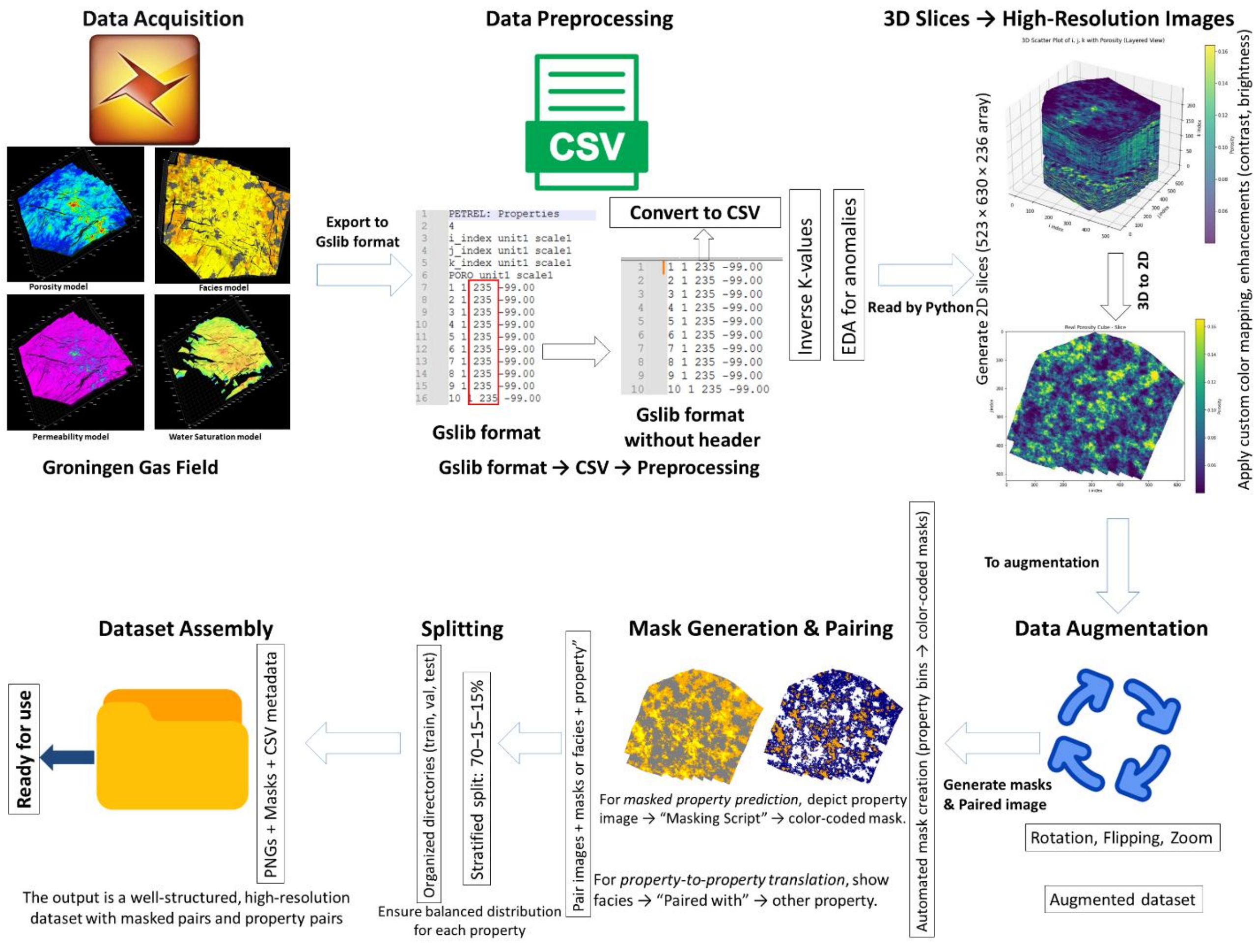

**Fig.1** Workflow schematic for the development of the GeoPix v1 dataset. The process begins with the extraction of geological property models from Petrel (porosity, permeability, water saturation, and independently interpreted facies), exported in GSLIB format. These are converted to CSV, cleaned, and reconstructed into 3D arrays from which 2D high-resolution slices are extracted. Custom color mapping and basic image enhancements are applied to each slice. Augmentation (rotation, flipping, zooming) expands the dataset, while automated mask generation enables semantic segmentation tasks. Final paired images are created for use in property-to-property translation and segmentation experiments. The dataset is stratified and organized into training, validation, and testing directories for downstream use.

A full summary of the final dataset structure, including the number of original slices, augmentation expansion, and stratified splits into training, validation, and testing sets, is provided in Table 1.

Table 1. GeoPix v1 dataset composition, augmentation, and dataset splits by property.

| Property | Original Slices | Augmentation Factor | Resulting Images | Training Set | Validation Set | Test Set |
|---|---|---|---|---|---|---|
| Facies | 235 slices | 10× | 2,350 | 1,809 | 387 | 389 |
| Porosity | 235 slices | 10× | 2,350 | 1,816 | 389 | 390 |
| Permeability | 235 slices | 10× | 2,350 | 1,809 | 387 | 389 |
| Sw | 235 slices | 10× | 2,350 | 1,694 | 363 | 363 |

To facilitate subsequent computational handling, large, high-precision GSLIB files (~1.6 GB per property) were converted into structured CSV format through a multi-step process (Figure 2):

- **Header removal**: non-data metadata lines (the first six lines of each file) were removed using Notepad++.
- **CSV Formatting**: Remaining data lines were standardized into consistent column structures (I, J, K, and property values).
- **Inverse Permeability Correction**: Permeability values initially recorded inversely were corrected by applying the transformation:

$$K_{corrected} = \max(k) - k + 1$$

This transformation ensured that the permeability values aligned with expected geological trends.

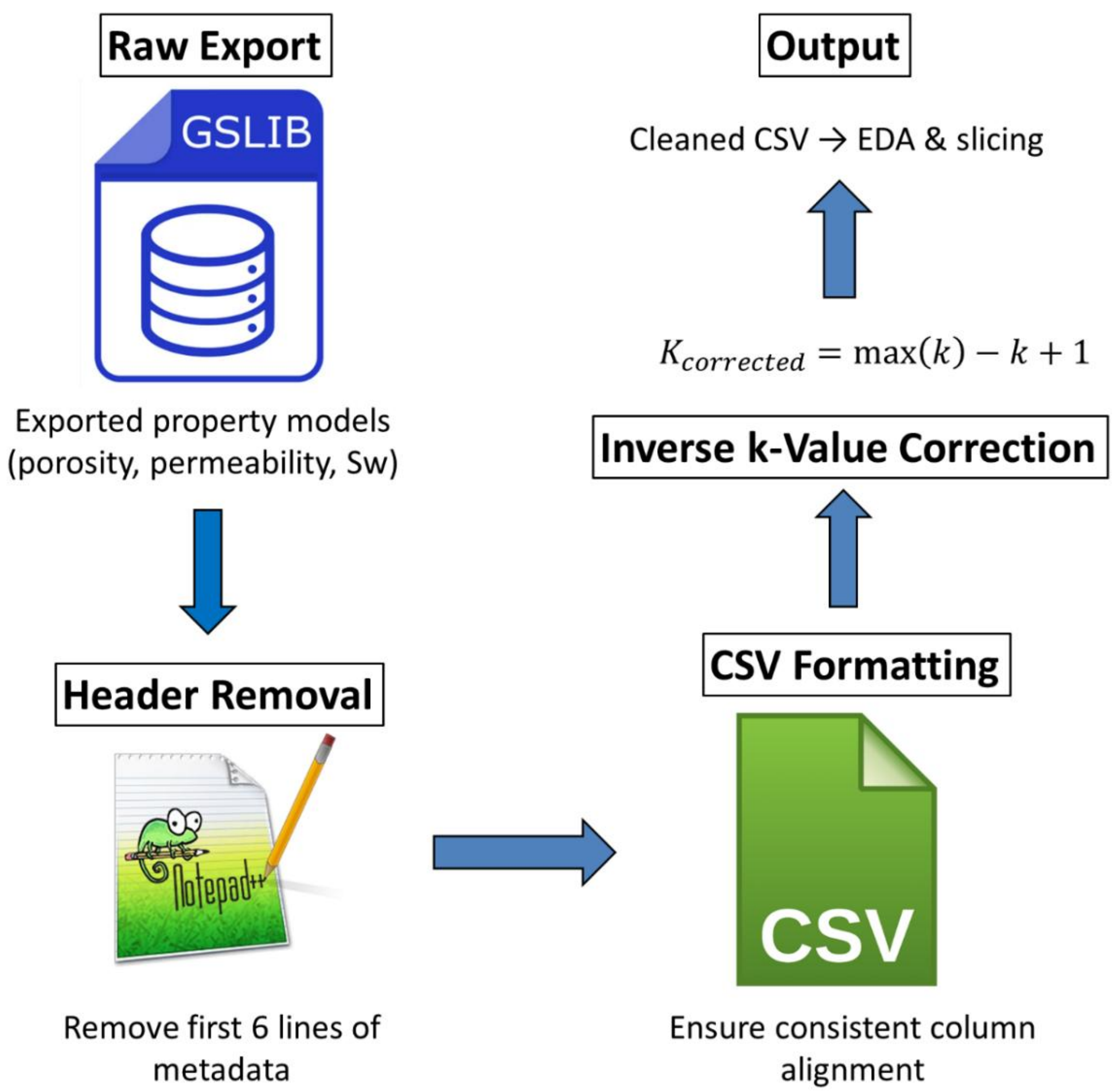

**Fig.2** Step-by-step conversion process of raw geological property exports from Petrel into cleaned CSV format. The workflow includes GSLIB export, metadata header removal, column alignment, and inverse permeability correction to produce standardized datasets suitable for 3D reconstruction and analysis.

Table 2 provides an overview of the file sizes, number of properties processed, and approximate time spent at each step, supplementing the more detailed transformation steps illustrated in Fig. 2.

Table 2. Summary of the key file sizes, number of properties, and approximate time spent during the GSLIB-to-CSV conversion and cleaning process.

| Step | File Size / Properties | Time Spent | Outcome |
|---|---|---|---|
| GSLIB Export | ~1.6 GB per property | 1–2 hours (per export) | Raw data with “Include cell index” enabled |

| Header Removal | 6 lines of non-data metadata | ~15–30 minutes | Cleaned text file, removing extraneous headers |
|---|---|---|---|
| CSV Formatting | 3–4 columns (I, J, K, property values) | ~30 minutes | Standardized CSV, ready for subsequent checks |
| Inverse k-Value Correction | Affects only permeability data | ~10–15 minutes | Corrected permeability range ($K_{corrected}$) |

**Data extraction and image generation**. After conversion to structured CSV format, the property data were reconstructed into three-dimensional arrays of 523 × 630 × 236 cells. Layer-wise extraction from these arrays enabled generation of aligned two-dimensional slices for each property domain. Fig. 3 summarizes the transformation from tabular property data to three-dimensional array representation and finally to two-dimensional colour-mapped slices. To preserve visual consistency with geological interpretation workflows, the slices were rendered using Petrel-like colour schemes and modest image enhancements, including contrast, brightness, and sharpening. The principal array-construction and rendering parameters are summarized in Table 3.

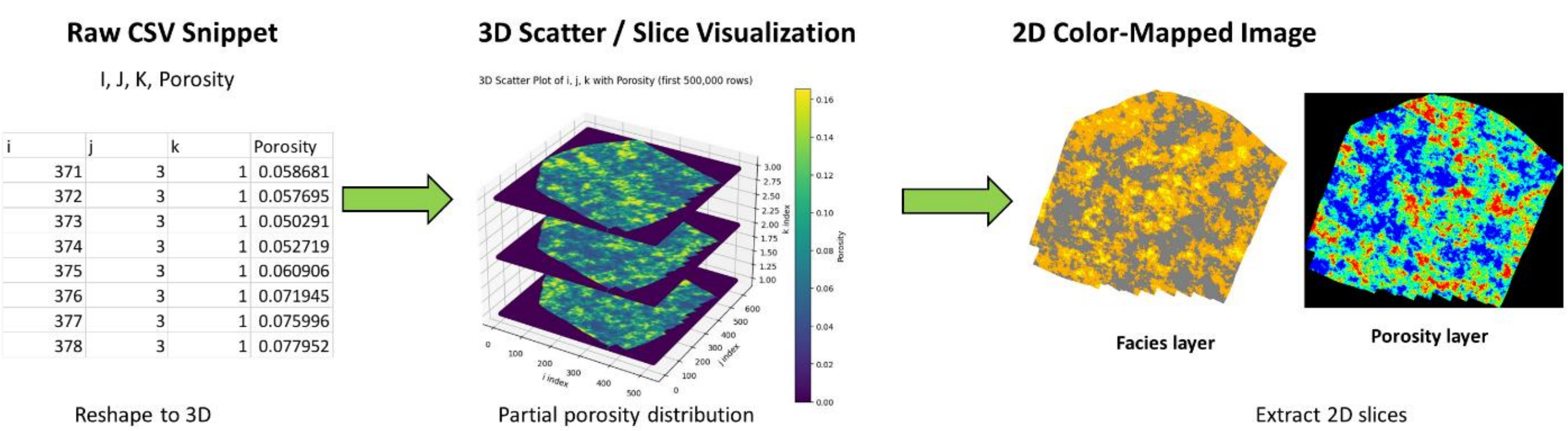

**Fig. 3** Transformation from raw CSV data (left) to a 3D volumetric representation (center) and ultimately to 2D color-mapped slices (right). The middle panel illustrates partial porosity distribution, while the final panel shows extracted 2D layers for facies and porosity, preserving critical geological variations for subsequent modeling.

Key parameters of 3D array construction and visualization are summarized in **Table 3**.

Table 3. Key parameters for the 3D array construction and subsequent color mapping of 2D reservoir slices

| Parameter | Value/Description |
|---|---|
| 3D Array Dimensions | 523 × 630 × 236 |
| Total Layers (K-direction) | 236 |
| Color Mapping Scheme | Petrel-like gradient (e.g., blue=low, red=high) |
| Image Enhancement | Contrast, brightness, and sharpening |
| Software & Libraries | Python (NumPy, Matplotlib), OpenCV |

**Data Augmentation.** To increase visual variability while preserving spatial structure, each of the 235 original two-dimensional slices per property was subjected to a reproducible augmentation workflow. TensorFlow ImageDataGenerator was used to apply rotations of ±10°, translations of up to 10% in width and height, shear transformations of ±10%, zoom of ±10%, horizontal flipping, and nearest-neighbour fill mode, yielding 10 augmented images per original slice. The augmentation parameters are summarized in Table 4, and representative examples are shown in Fig. 4. These augmented products are not part of the fixed deposited dataset and are instead regenerated through the archived software workflow.

Table 4. Summary of the data augmentation techniques and parameter values used in GeoPix v1.

| Augmentation Technique | Parameter Value |
|---|---|
| Rotation | ±10° |
| Width/Height shift | ±10% |
| Shear | ±10% |
| Zoom | ±10% |
| Horizontal flip | Yes |
| Fill mode | Nearest |
| Augmentation factor | 10× per image |
| Original images | 235 per property |
| Total augmented images | 2,350 per property |

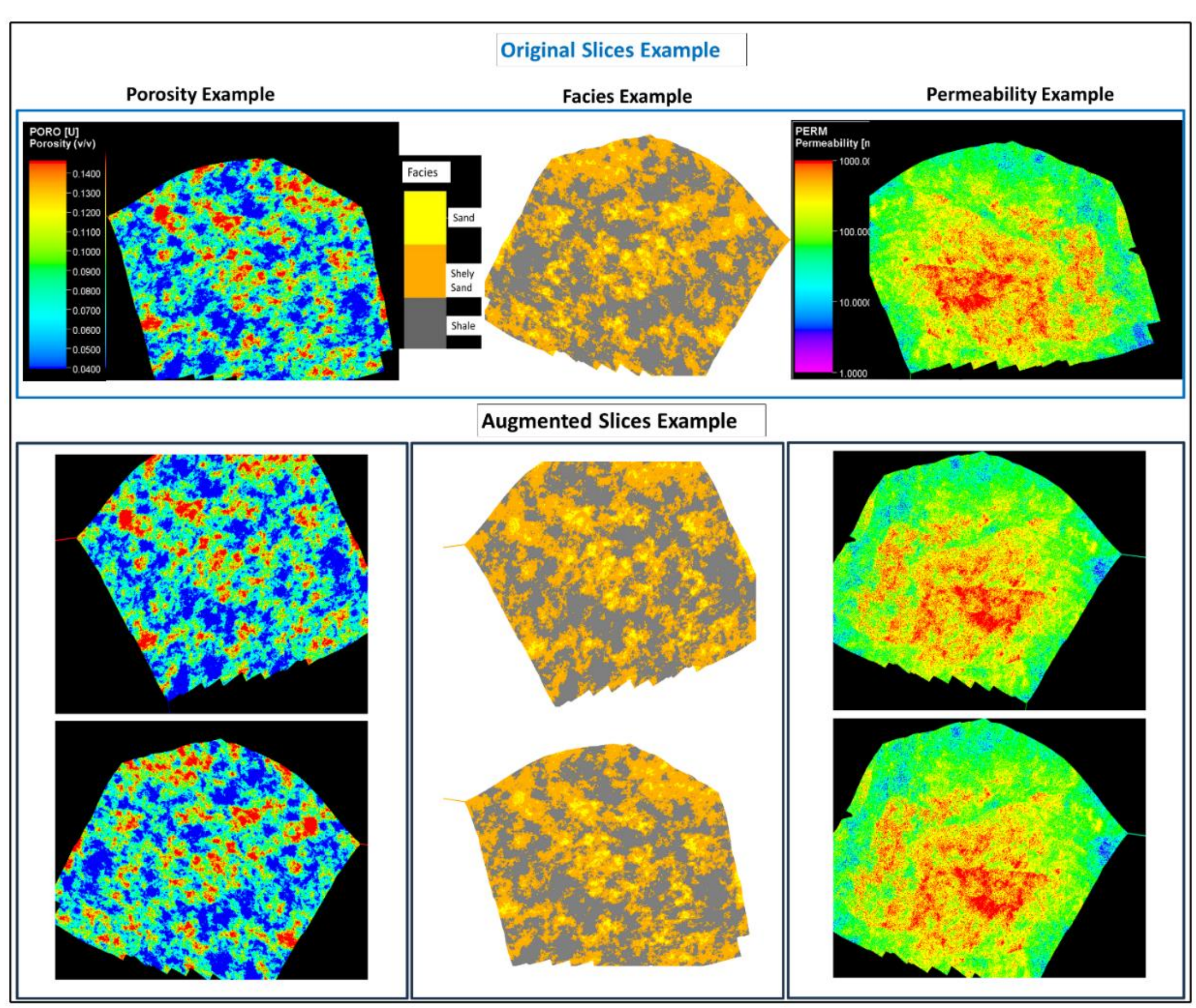

**Fig. 4** Examples of original and augmented 2D slices from the GeoPix dataset. The top row displays original slices for porosity, facies, and permeability extracted from the Groningen geological model. The bottom rows show corresponding augmented slices generated using geometric transformations (rotation, flipping, shifting, and zooming). These augmentations enhance model generalization by increasing the visual diversity of spatial patterns without requiring additional labeled data.

Augmented images were saved systematically using a structured naming convention (e.g., slice001_aug01.png) and merged with the original images into a comprehensive dataset suitable for modeling tasks.

**Mask generation and image pairing.** In addition to the original image corpus, the archived software workflow supports generation of auxiliary semantic masks and paired images for supervised learning tasks[40]. Mask generation was based on image-intensity thresholds applied to the rendered slices, producing four image-derived classes summarized in Table 5. Paired images were then assembled for two downstream uses: property–mask pairing for segmentation experiments and facies–property pairing for supervised image-to-image translation. Fig. 5 presents representative examples of paired images and their corresponding derived masks. These products are reproducible from the archived software workflow and are described here as secondary outputs rather than as part of the fixed deposited dataset.

**Table 5.** Semantic classification of pixel intensity ranges used to generate automated property masks in GeoPix v1, supporting segmentation and supervised image-pairing tasks.

| Mask Color | Geological Feature | Intensity Threshold |
|---|---|---|
| Yellow | High geological significance | >200 (bright) |
| Orange | Intermediate geological features | >150 |
| White | Homogeneous geological background | >100 |
| Dark Blue | Non-geological background | <100 |

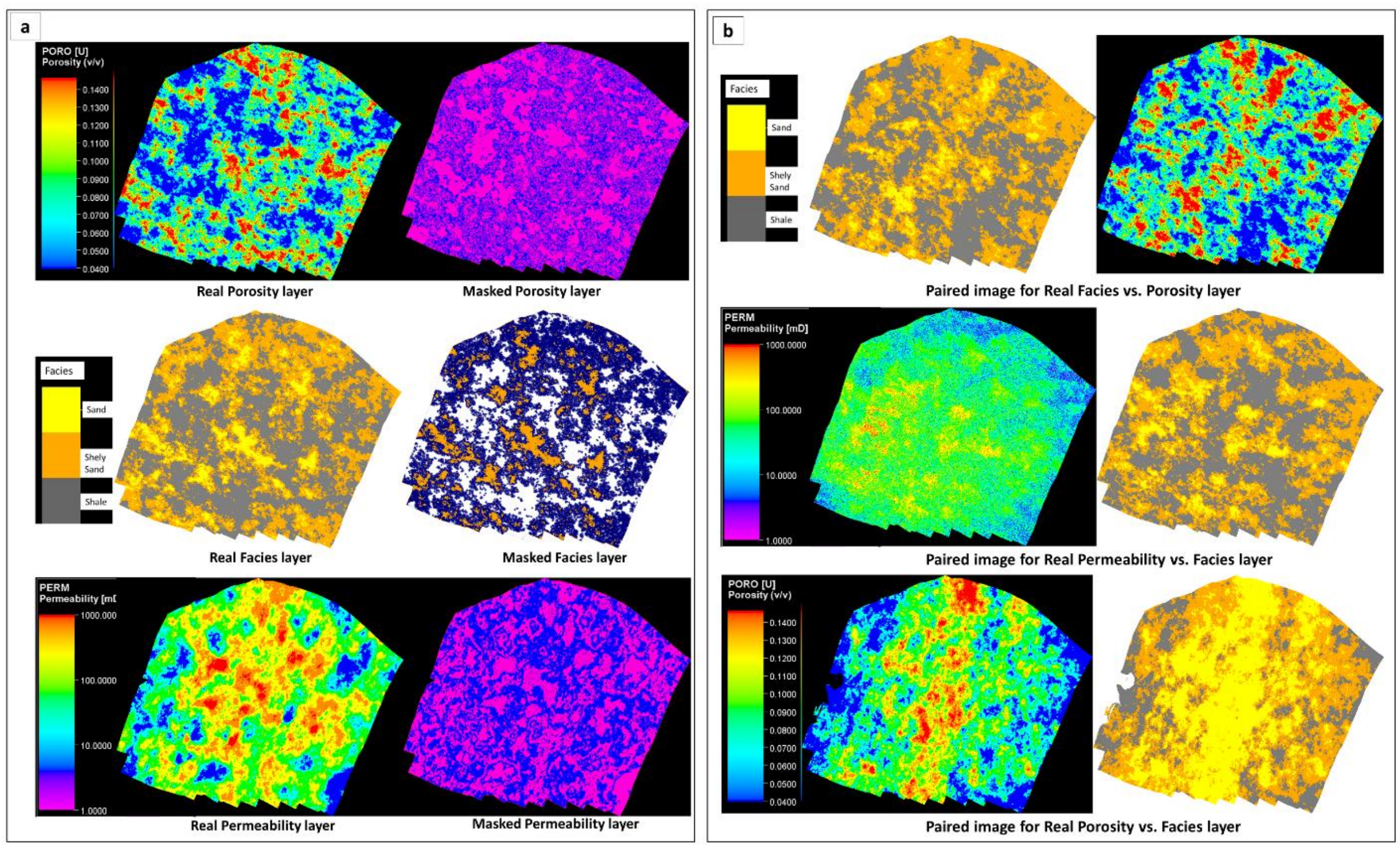

**Fig. 5** Representative examples of paired images and automatically generated semantic masks for the GeoPix dataset. Original augmented images (left side of each pair) are presented alongside their corresponding semantic masks (right side), highlighting clear differentiation of geological features suitable for Pix2Pix cGAN-based modeling tasks.

**Computational Environment.** All computational processing, data handling, augmentation, and baseline model training, was performed using a high-performance workstation equipped with:

- **Hardware**: 4× NVIDIA RTX A5500 GPUs (24 GB VRAM each), 128 GB RAM, and a multi-core CPU optimized for parallel computing.
- **Software & Libraries**: Python 3.8; TensorFlow and Keras (for data augmentation using ImageDataGenerator); PyTorch (for sample cGAN training); and OpenCV and Matplotlib (for visualization and image handling).

This configuration enabled efficient processing of high-resolution 2D and 3D data throughout the GeoPix pipeline. Fig.6 illustrates the complete hardware and software stack used during dataset preparation and model prototyping.

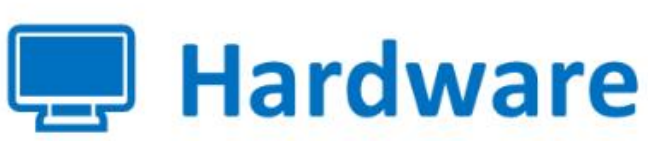

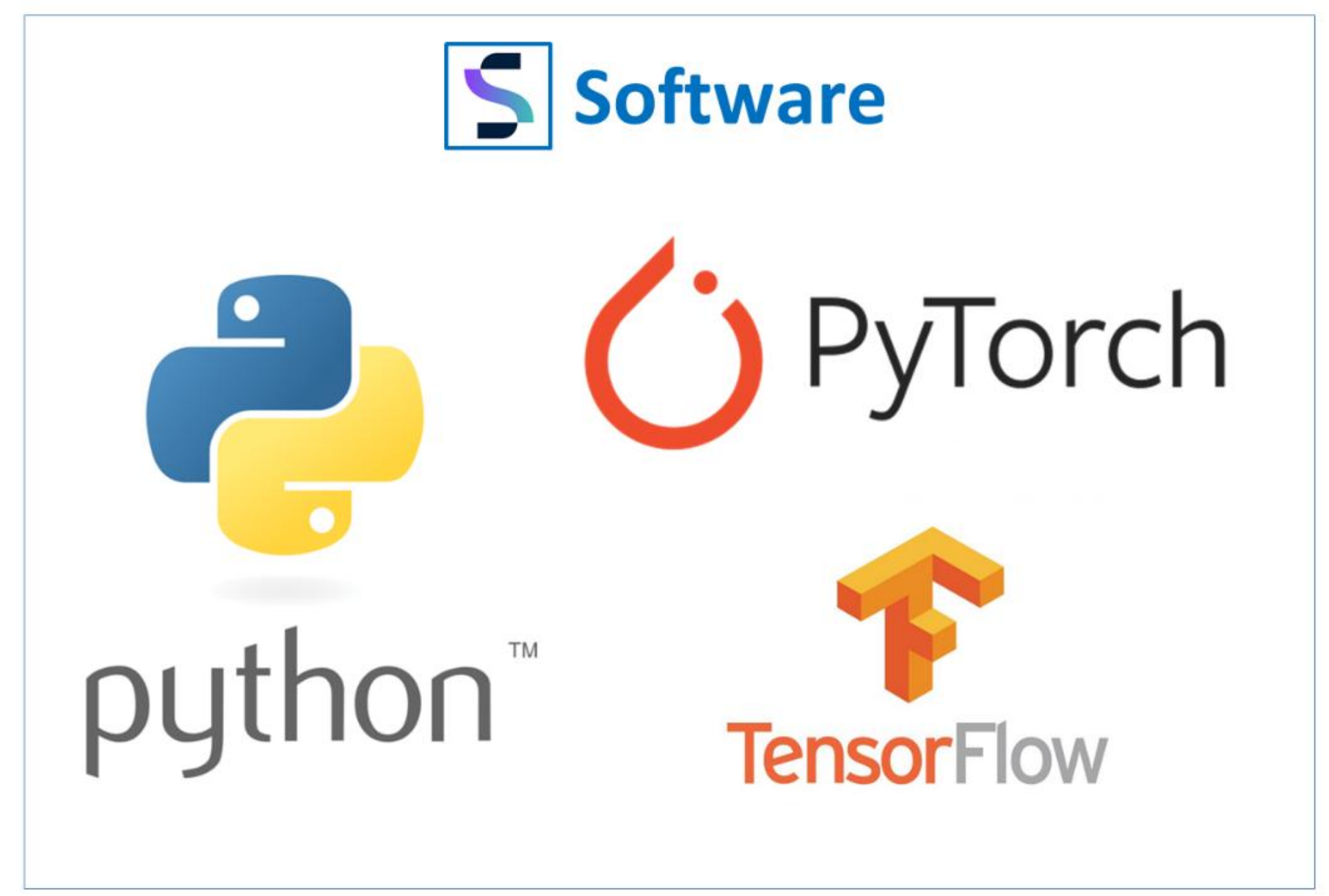

**Fig.6** Overview of the hardware and software environment used for dataset preparation, augmentation, and model training. The pipeline was executed on a high-performance workstation with four NVIDIA RTX A5500 GPUs and 128 GB RAM, leveraging Python-based deep learning libraries including TensorFlow and PyTorch.

## Data Records

The GeoPix v1 resource is distributed through two linked Zenodo records. The original high-resolution reservoir-property images are deposited as a dataset at Zenodo[1]. The code used to reproduce augmentation, mask generation, paired-image construction, and example baseline workflows is archived as software at Zenodo[2] and is also available through the project GitHub repository for active access and version tracking.

The dataset record[1] contains the original PNG image slices extracted from the Groningen static reservoir model for four domains: facies, porosity, permeability, and water saturation (Sw). These images preserve the spatial resolution and visual styling used during export and constitute the fixed primary data product described in this paper.

The software record[2] contains the scripts and notebooks required to reproduce the derived products built from the original images. These derived products include:

(1) augmented image sets generated through geometric transformations;

(2) semantic masks created from image-based classification rules; and

(3) paired images assembled for supervised image-to-image translation tasks such as facies-to-porosity, facies-to-permeability, facies-to-Sw, and related mappings.

The dataset is organized to support two main reuse scenarios. The first is direct use of the original exported images for visualization, benchmarking, and custom preprocessing. The second is reproducible generation of task-specific machine-learning inputs using the archived software workflow[2]. This modular structure keeps the deposited dataset compact while allowing users to regenerate extended products in a transparent and version-controlled manner.

The original dataset record[1] contains the four image folders corresponding to facies, porosity, permeability, and Sw. Within each folder, images are stored as layer-based PNG files using a consistent naming structure. The software archive[2] provides the processing scripts, example outputs, and supporting documentation required to reconstruct augmentation products, masks, paired images, and baseline demonstrations. Together, these records define the complete GeoPix v1 resource.

## Technical Validation

The technical quality of the GeoPix resource was assessed through checks on image integrity, cross-domain correspondence, and workflow reproducibility. The deposited image corpus was reviewed for dimensional consistency, naming consistency, and slice-level correspondence among facies, porosity, permeability, and water saturation. Representative subsets of the original and regenerated derived products were visually inspected to verify colour consistency, alignment, and preservation of structural patterns. In addition, the archived workflow was used to reproduce example Pix2Pix-based translation experiments as a practical demonstration that the dataset supports supervised cross-domain learning and can be reused in a reproducible manner.

## Usage Notes

GeoPix is intended for reuse in reservoir-property visualization, semantic segmentation, and supervised image-to-image translation workflows. Users interested only in the fixed image corpus can work directly with the Zenodo dataset record. Users requiring augmented images, semantic masks, paired images, or baseline examples should regenerate these products from the archived software workflow. Because the dataset is distributed as rendered image slices, users should distinguish between the deposited visual representations and the underlying physical property values represented by the source model. Reuse involving quantitative physical interpretation should therefore refer back to the original Groningen source publication and the documented export settings.

## Data Availability

The original high-resolution GeoPix v1 image dataset is available at Zenodo[1]. The archived GeoPix processing workflow, including code for augmentation, mask generation, paired-image construction, and example baseline demonstrations, is available at Zenodo[2]. The software is also mirrored through the project GitHub repository for active access and updates. Repository contents and file organization are described in the Data Records section.

## Acknowledgments

The authors acknowledge King Fahd University of Petroleum and Minerals for institutional support. The authors also thank Nederlandse Aardolie Maatschappij, EPOS-NL, and Utrecht University for enabling public access to the Groningen source model used in this work.

## Funding

This research received no external funding.

## Author Contributions

A.-F. conceived the dataset, developed the processing workflow, curated the data, performed the analyses, prepared the figures, and wrote the original draft. A.K. contributed to supervision, resources, and manuscript review. N.A.S. contributed to visualization, manuscript structuring, and manuscript review. S.I.K. contributed to supervision, project administration, and manuscript review.

## Competing Interests

The authors declare no competing interests.

## Code Availability

Custom code used for preprocessing, augmentation, mask creation, image pairing, and example Pix2Pix-based workflows is archived at Zenodo[2]. The same codebase is also available through the project GitHub repository for active development and maintenance. The archived Zenodo software record represents the fixed version associated with this manuscript.